\documentclass[11pt,a4paper]{article}
\usepackage[hyperref]{acl2021}
\usepackage{times}
\usepackage{latexsym}

\usepackage{microtype}

\aclfinalcopy % Uncomment this line for the final submission
\usepackage{tabularx}
\usepackage{subfig}
\usepackage{graphicx}
\usepackage{bm}
\usepackage{amssymb}
\usepackage{amsmath}
\usepackage{xcolor}
\usepackage{multirow}
\usepackage{booktabs}
\usepackage{verbatim}
\usepackage{mdwlist}
\usepackage[ruled,norelsize,linesnumbered]{algorithm2e}
\usepackage{amsthm}
\usepackage[capitalize]{cleveref}
\usepackage{paralist}
\usepackage{enumitem}

\makeatletter
\newcommand{\removelatexerror}{\let\@latex@error\@gobble}
\makeatother

\usepackage{amsmath,amsfonts,bm}

\def\eqref#1{equation~\ref{#1}}
\def\1{\bm{1}}

\DeclareMathAlphabet{\mathsfit}{\encodingdefault}{\sfdefault}{m}{sl}
\SetMathAlphabet{\mathsfit}{bold}{\encodingdefault}{\sfdefault}{bx}{n}

\newcommand{\sigmoid}{\sigma}

\title{Soft Layer Selection with Meta-Learning for Zero-Shot\\Cross-Lingual Transfer}

\author{Weijia Xu\thanks{~~Work done while interning at Amazon AI.}$~\dagger~~$ 
  Batool Haider$\ddagger~~$
  Jason Krone$\ddagger~~$
  Saab Mansour$\ddagger~~$ \\
  $\dagger$Department of Computer Science, University of Maryland \\
  $\ddagger$Amazon AI  \\
  {\tt weijia@cs.umd.edu, \{bhaider, kronej, saabm\}@amazon.com} \\
  }

\date{}

\begin{document}
\maketitle
\begin{abstract}
Multilingual pre-trained contextual embedding models~\citep{Devlin2019} have achieved impressive performance on zero-shot cross-lingual transfer tasks. Finding the most effective strategy to fine-tune these models on high-resource languages so that it transfers well to the zero-shot languages is a non-trivial task. In this paper, we propose a novel meta-optimizer to soft-select which layers of the pre-trained model to freeze during fine-tuning. We train the meta-optimizer by simulating the zero-shot transfer scenario.
Results on cross-lingual natural language inference show that our approach improves over the simple fine-tuning baseline and X-MAML~\citep{NooralahzadehBBA2020}.
\end{abstract}

\section{Introduction}

Despite the impressive performance of neural models on a wide variety of NLP tasks, these models are extremely data hungry \---\ training them requires a large amount of annotated data. As collecting such amounts of data for every language of interest is extremely expensive, cross-lingual transfer that aims to transfer the task knowledge from high-resource~(\textit{source}) languages for which annotated data are more readily available to low-resource~(\textit{target}) languages becomes a promising direction.
Cross-lingual transfer approaches using cross-lingual resources such as machine translation~(MT) systems~\citep{Wan2009,Conneau2018} or bilingual dictionaries~\citep{Prettenhofer2010} have effectively reduced the amount of annotated data required to obtain reasonable performance on the target language. However, such cross-lingual resources are often limited for low-resource languages.

\looseness=-1
Recent advances in cross-lingual contextual embedding models have reduced the need for cross-lingual supervision~\citep{Devlin2019,LampleC2019}.
\citet{WuD2019} show that multilingual BERT~(mBERT)~\citep{Devlin2019}, a contextual embedding model pre-trained on the concatenated Wikipedia data from~104 languages without cross-lingual alignment, does surprisingly well on zero-shot cross-lingual transfer tasks, where they fine-tune the model on the annotated data from the source languages and evaluate on the target language. \citet{WuD2019} propose to freeze the bottom layers of mBERT during fine-tuning to improve the cross-lingual performance over the simple fine-tune-all-parameters strategy, as different layers of mBERT captures different linguistic information~\citep{Jawahar2019}.

\looseness=-1
Selecting which layers to freeze for a downstream task is a non-trivial problem. In this paper, we propose a novel meta-learning algorithm for soft layer selection. 
Our meta-learning algorithm learns layer-wise update rate by simulating the zero-shot transfer scenario \---\ at each round, we randomly split the source languages into a held-out language and the rest as training languages, fine-tune the model on the training languages, and update the meta-parameters based on the model performance on the held-out language. We build the meta-optimizer on top of a standard optimizer and learnable update rates, so that it generalizes well to large numbers of updates. Our method uses much less meta-parameters than the X-MAML approach~\citep{NooralahzadehBBA2020} adapted from model-agnostic meta-learning~(MAML)~\citep{Finn2017} to zero-shot cross-lingual transfer.

\looseness=-1
Experiments on zero-shot cross-lingual natural language inference show that our approach outperforms both the simple fine-tuning baseline and the X-MAML algorithm and that our approach brings larger gains when transferring from multiple source languages. Ablation study shows that both the layer-wise update rate and cross-lingual meta-training are key to the success of our approach.

\section{Meta-Learning for Zero-Shot Cross-lingual Transfer}
\label{sec:model}

\looseness=-1
The idea of transfer learning is to improve the performance on the target task~$\mathcal{T}^0$ by learning from a set of related source tasks~$\{\mathcal{T}^1, \mathcal{T}^2, ..., \mathcal{T}^K\}$. In the context of cross-lingual transfer, we treat different languages as separate tasks, and our goal is to transfer the task knowledge from the source languages to the target language. In contrast to the transfer learning case where the inputs of the source and target tasks are from the same language, in cross-lingual transfer learning we need to handle inputs from different languages with different vocabularies and syntactic structures. To handle the issue, we use the pre-trained multilingual BERT~\citep{Devlin2019}, a language model encoder trained on the concatenation of monolingual corpora from~104 languages.

The most widely used approach to zero-shot cross-lingual transfer using multilingual BERT is to fine-tune the BERT model~$\theta$ on the source language tasks~$\mathcal{T}^{1...K}$ with training objective~$\mathcal{L}$
\[
    \boldsymbol{\theta}^* = \text{Learn}(\mathcal{L}, \mathcal{T}^1, ..., \mathcal{T}^K; \boldsymbol{\theta})
\]
and then evaluate the fine-tuned model~$\boldsymbol{\theta}^*$ on the target language task~$\mathcal{T}^0$. The gap between training and testing can lead to sub-optimal performance on the target language.

To address the issue, we propose to train a meta-optimizer~$f_\phi$ for fine-tuning so that the fine-tuned model generalizes better to unseen languages. We train the meta-optimizer by
\[
    \boldsymbol{\phi}^* = \text{Learn}(\mathcal{L}, \mathcal{T}^k; \text{MetaLearn}(\mathcal{L}, \mathcal{T}^{1...K} \setminus \mathcal{T}^k; \boldsymbol{\phi}))
\]
where~$\mathcal{T}^k$ is a ``surprise'' language randomly selected from the source language tasks~$\mathcal{T}^{1...K}$.

\begin{figure}[!ht]
\removelatexerror
\begin{algorithm}[H]
\label{alg:train}
\SetAlgoLined
\KwIn{
    Training data~$\{\mathcal{D}_1, ..., \mathcal{D}_K\}$ in the source languages, learner model~$M$ with parameters~$\boldsymbol{\theta}$, and meta-optimizer with base optimizer~$f_{opt}$ and meta-parameters~$\boldsymbol{\phi}$.
}
\KwOut{
    Meta-optimizer with parameters~$\boldsymbol{\phi}$.
}
\BlankLine
$s \leftarrow 1$ \\
Randomly initialize~$\boldsymbol{\phi}^0$. \\
\BlankLine
\SetKwFor{RepTimes}{repeat}{times}{end}
\RepTimes{$N$} {
%\For{$i \gets 1$ \KwTo $N$} {
    \BlankLine
    $t \leftarrow 1$ \\
    Initialize~$\boldsymbol{\theta}^0$ with mBERT and random values for the classification layer. \\
    Randomly select a test language~$k$ to form the test data~$\mathcal{D}_{test} = \mathcal{D}_{k}$. \\
    $\mathcal{D}_{train} \leftarrow \{\mathcal{D}_1, ..., \mathcal{D}_K\} \setminus \mathcal{D}_{test}$ \\
    \BlankLine
    \RepTimes{$L$} {
    %\For{$d \gets 1$ \KwTo $L$} {
        \BlankLine
        $\boldsymbol{X}^t, \boldsymbol{Y}^t$ $\leftarrow$ random batch from~$\mathcal{D}_{train}$ \\
        $\mathcal{L}^t \leftarrow \mathcal{L}(M(\boldsymbol{X}^t; \boldsymbol{\theta}^{t-1}), \boldsymbol{Y}^t)$ \\
        $\boldsymbol{g}^{1...t} \leftarrow [\boldsymbol{g}^{1...t-1}, \nabla_{\boldsymbol{\theta}^{t-1}}\mathcal{L}^t]$ \\
        $\Delta \boldsymbol{\theta}^t \leftarrow f_{opt}(\boldsymbol{g}^1, ..., \boldsymbol{g}^t)$ \\
        $\boldsymbol{\theta}^t \leftarrow \boldsymbol{\theta}^{t-1} - \sigmoid(\boldsymbol{\phi}^{s-1}) \odot \Delta \boldsymbol{\theta}_t$ \\
        $t \leftarrow t + 1$ \\
    }
    \BlankLine
    $\boldsymbol{X}, \boldsymbol{Y} \leftarrow \mathcal{D}_{test}$ \\
    $\mathcal{L}_{test} \leftarrow \mathcal{L}(M(\boldsymbol{X}; \boldsymbol{\theta}^t), \boldsymbol{Y})$ \\
    $\boldsymbol{\phi}^s \leftarrow \text{Update}(\boldsymbol{\phi}^{s-1}, \nabla_{\boldsymbol{\phi}^{s-1}}\mathcal{L}_{test})$ \\
    $s \leftarrow s + 1$ \\
}
\caption{Meta-Training}
\end{algorithm}
\end{figure}

\begin{figure*}[ht]
    \centering
    \includegraphics[width=0.85\textwidth]{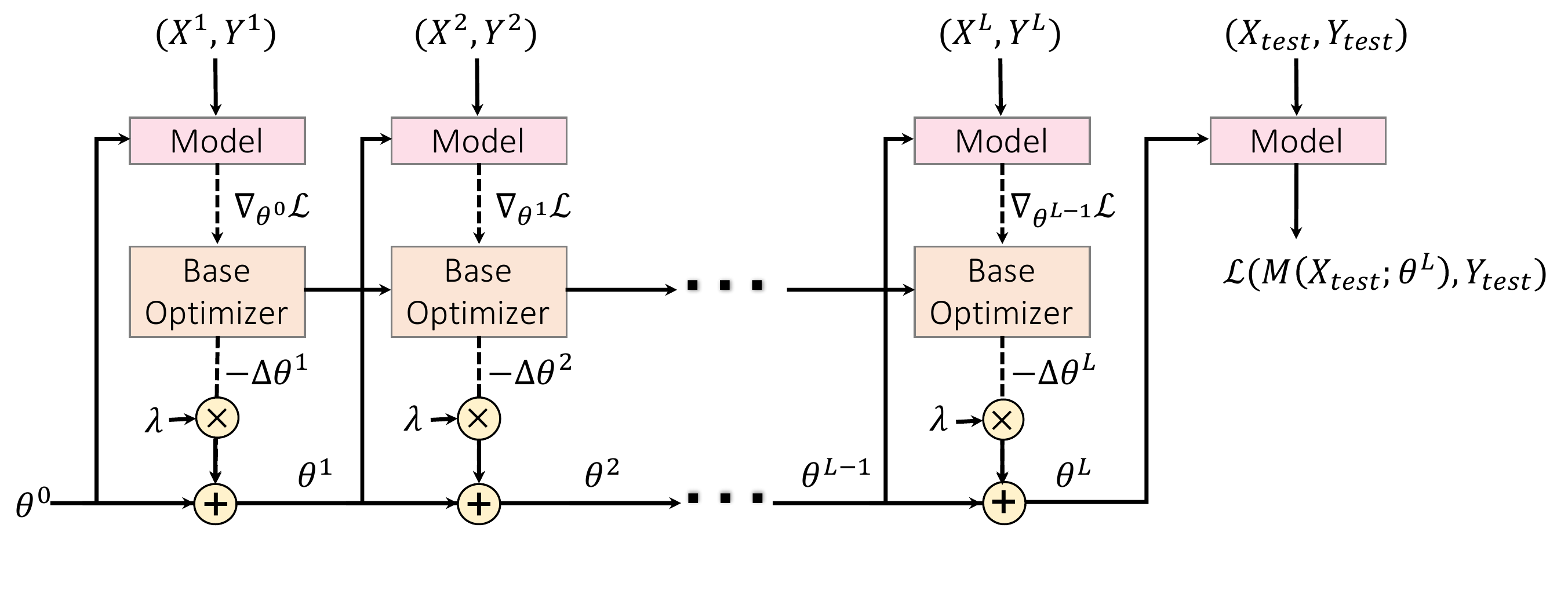}
\caption{Computational graph for the forward pass of the meta-optimizer. Each batch~$(\boldsymbol{X}^t, \boldsymbol{Y}^t)$ is from the training data~$\mathcal{D}_{train}$, and~$(\boldsymbol{X}_{test}, \boldsymbol{Y}_{test})$ denotes the entire test set. The meta-learner is comprised of a base optimizer that takes the history and current step gradients as inputs and suggests an update~$\Delta \boldsymbol{\theta}^t$, and the meta parameters that control the layer-wise update rates~$\boldsymbol{\lambda}$ for the learner model~$\boldsymbol{\theta}$. The dashed arrows indicate that we do not back-propagate the gradients through that step when updating the meta-parameters.}
\label{fig:meta_optimizer}
\end{figure*}

\subsection{Meta-Optimizer}
Our meta-optimizer consists of a standard optimizer as the base optimizer and a set of meta-parameters to control the layer-wise update rates. An update step is formulated as:
\begin{equation}
\begin{split}
    \boldsymbol{\theta}^{t} = \boldsymbol{\theta}^{t-1} - \boldsymbol{\lambda} \odot \Delta \boldsymbol{\theta}^t \\
    \Delta \boldsymbol{\theta}^t = f_{opt}(\boldsymbol{g}^1, ..., \boldsymbol{g}^t)
\end{split}
\label{eq:upd}
\end{equation}
where~$\boldsymbol{\theta}^t$ represent the parameters of the learner model at time step~$t$, and~$\Delta \boldsymbol{\theta}^t$ is the update vector produced by the base optimizer~$f_{opt}$ given the gradients~$\{\boldsymbol{g}^i = \nabla_{\boldsymbol{\theta}^{i-1}}\mathcal{L}^i\}_{i=1}^t$ at the current and previous steps.
The function~$f_{opt}$ is defined by the optimization algorithm and its hyper-parameters. For example, a typical gradient descent algorithm uses~$f_{opt} = \alpha \boldsymbol{g}^t$ where~$\alpha$ represents the learning rate. A standard optimization algorithm will update the model parameters by:
\begin{equation}
    \boldsymbol{\theta}^t = \boldsymbol{\theta}^{t-1} - f_{opt}(\boldsymbol{g}^1, ..., \boldsymbol{g}^t)
\end{equation}
Our meta-optimizer is different in that we perform gated update using parametric update rates~$\boldsymbol{\lambda}$, which is computed by~$\boldsymbol{\lambda} = \sigmoid(\boldsymbol{\phi})$, where~$\boldsymbol{\phi}$ represents the meta-parameters of the meta-optimizer~$f_\phi$. The sigmoid function ensures that the update rates are within the range~$[0, 1]$.
Different from~\citet{Andrychowicz2016} in which the optimizer parameters are shared across all coordinates of the model, our meta-optimizer learns different update rates for different model layers. This is based on the findings that different layers of the BERT encoder capture different linguistic information, with syntactic features in middle layers and semantic information in higher layers~\citep{Jawahar2019}. And thus, different layers may generalize differently across languages.

\looseness=-1
Figure~\ref{fig:meta_optimizer} illustrates the computational graph for the forward pass when training the meta-optimizer. Note that as the losses~$\mathcal{L}^t$ and gradients~$\nabla_{\boldsymbol{\theta}^{t-1}}\mathcal{L}^t$ are dependent on the parameters of the meta-optimizer, computing the gradients along the dashed edges would normally require taking second derivatives, which is computationally expensive. Following~\citet{Andrychowicz2016}, we drop the gradients along the dashed edges and only compute gradients along the solid edges.

\subsection{Meta-Training}

\looseness=-1
A good meta-optimizer will, given the training data in the source languages and the training objective, suggest an update rule for the learner model so that it performs well on the target language. Thus, we would like the training condition to match that of the test time. However, in zero-shot transfer we assume no access to the target language data, so we need to simulate the test scenario using only the training data on the source languages.

\looseness=-1
As shown in Algorithm~\ref{alg:train}, at each episode in the outer loop, we randomly choose a test language~$k$ to construct the test data~$\mathcal{D}_{test} = \mathcal{D}_{k}$ and use the remaining data as the training data~$\mathcal{D}_{train}$. Then, we re-initialize the parameters of the learner model and start the training simulation.
At each training step, we first use the base optimizer~$f_{opt}$ to compute the update vector~$\Delta \boldsymbol{\theta}^t$ based on the current and history gradients~$\boldsymbol{g}^{1...t}$. We then perform the gated update using the meta-optimizer~$\boldsymbol{\phi}^{s-1}$ with~\cref{eq:upd}. The resulting model~$\boldsymbol{\theta}^t$ can be viewed as the output of a forward pass of the meta-optimizer.
After every~$L$ iterations of model update, we compute the gradient of the loss on the test data~$\mathcal{D}_{test}$ with respect to the old meta parameters~$\boldsymbol{\phi}^{s-1}$ and make an update to the meta parameters.
Our meta-learning algorithm is different from X-MAML~\citep{NooralahzadehBBA2020} in that
\begin{inparaenum}[1)]
    \item X-MAML is designed mainly for few-shot transfer while our algorithm is designated for zero-shot transfer, and
    \item our algorithm uses much less meta-parameters than X-MAML as it only requires training the update rate for each layer while in X-MAML we meta-learn the initial parameters of the entire model.
\end{inparaenum}
\section{Experiments}
\label{sec:exp}

\renewcommand{\arraystretch}{1.2}
\begin{table*}
\centering
\small
\setlength{\tabcolsep}{0.2em}
\scalebox{0.95}{
\begin{tabular}{lccccccccccccccc}
\toprule
& fr &	es &	de &	ar &	ur &	bg &	sw &    th &	tr &	vi & 	zh & 	ru &	el &	hi &    avg \\
\midrule
{\citet{Devlin2019}} & {--} & {74.30} & {70.50} & {62.10} & {58.35} & {--} & {--} & {--} & {--} & {--} & {63.80} & {--} & {--} & {--} & {--} \\
\citet{WuD2019} & 74.60 &	74.90 &	72.00 &	66.10 &	58.60 &	69.80 &	49.40 &	55.70 &	62.00 &	71.90 &	70.40 &	69.80 &	67.90 &	61.20 &	66.02 \\
\citet{NooralahzadehBBA2020} & 74.42 &	75.07 &	71.83 &	66.05 &	61.51 &	69.45 &	49.76 &	55.39 &	61.20 &	71.82 &	71.11 &	70.19 &	67.95 &	62.20 &	66.28 \\
\midrule
Aux. language & el &	el & 	el &	el &	el &	el &	el &	el &	el &	el &	ur &	ur &	ur &	ur \\
Fine-tuning baseline & 75.42 &	75.77 &	72.57 & 67.22 &	61.08 &	70.23 &	\textbf{51.70} &	\textbf{51.03} &	\textbf{64.26} &	71.61 &	\textbf{72.52} &	69.97 &	69.16 &	55.40 &	66.28 \\
Meta-Optimizer & \textbf{75.78} &	\textbf{75.87} &	\textbf{73.15} &	\textbf{67.34} &	\textbf{62.00} &	\textbf{70.47} &	51.22 &	50.54 &	63.96 &	\textbf{72.06} &	72.32 &	\textbf{70.20} &	\textbf{69.34} &	\textbf{55.88} &	\textbf{66.44} \\
\midrule
\multicolumn{3}{l}{Aux. language: el + ur} \\
Fine-tuning baseline & 74.87 &    75.78 &	72.27 &	66.96 &	62.73 &	70.16 &	50.21 &	48.20 &	63.86 &	71.61 &	71.97	 & 70.24 &	69.64 &	56.04 &	66.04 \\
Meta-Optimizer& \textbf{75.53} &	\textbf{75.93} &	\textbf{72.68} &	\textbf{67.04} &	\textbf{63.33} &	\textbf{70.88} &	\textbf{51.51} &	\textbf{49.89} &	\textbf{64.33} &	\textbf{72.06} &	\textbf{72.36} &	\textbf{70.32} &	\textbf{70.38} &	\textbf{56.29} &	\textbf{66.61} \\
\bottomrule
\end{tabular}}
\caption{Accuracy of our approach compared with baselines on the XNLI dataset~(averaged over five runs). We compare our approach~(\textit{Meta-Optimizer}) with our fine-tuning baseline with one or two auxiliary languages, the fine-tuning results in \citet{Devlin2019}, the highest scores~(with a selected subset of layers fixed during fine-tuning) in \citet{WuD2019}, the best zero-shot results using X-MAML~\citep{NooralahzadehBBA2020} with one auxiliary language. We boldface the highest scores within each auxiliary language setting.}
\label{tab:xnli_result}
\end{table*}

\looseness=-1
We evaluate our meta-learning approach on natural language inference.
Natural Language Inference (NLI) can be cast into a sequence pair classification problem where, given a premise and a hypothesis sentence, the model needs to predict whether the premise entails the hypothesis, contradicts it, or neither~(neutral). We use the Multi-Genre Natural Language Inference Corpus~\citep{Williams2018}, which consists of~433k English sentence pairs labeled with textual entailment information, and the XNLI dataset~\citep{Conneau2018}, which has~2.5k development and~5k test sentence pairs in~15 languages including English~(en), French~(fr), Spanish~(es), German~(de), Greek~(el), Bulgarian~(bg), Russian~(ru), Turkish~(tr), Arabic~(ar), Vietnamese~(vi), Thai~(th), Chinese~(zh), Hindi~(hi), Swahili~(sw), and Urdu~(ur). We use this dataset to evaluate the effectiveness of our meta-learning algorithm when transferring from English and one or more low-resource auxiliary languages to the target language.

\renewcommand{\arraystretch}{1.2}
\begin{table*}[ht]
\centering
\small
\setlength{\tabcolsep}{0.2em}
\begin{tabular}{lccccccccccccccc}
\toprule
& fr &	es &	de &	ar &	ur &	bg &	sw &    th &	tr &	vi & 	zh & 	ru &	el &	hi &    avg \\
\midrule
Meta-Optim & \textbf{75.53} &	\textbf{75.93} &	\textbf{72.68} &	\textbf{67.04} &	\textbf{63.33} &	\textbf{70.88} &	\textbf{51.51} &	\textbf{49.89} &	\textbf{64.33} &	\textbf{72.06} &	\textbf{72.36} &	\textbf{70.32} &	\textbf{70.38} &	\textbf{56.29} &	\textbf{66.61} \\
{No layer-wise update} & 73.45 &	73.90 &	70.73 &	65.19 &	60.31 &	69.10 &	50.87 &	46.47 &	62.74 &	70.42 &	70.24 &	68.85 &	68.17 &	53.50 &	64.57 \\
{No cross-lingual meta-train} & 73.66 &	74.84 &	71.54 &	66.15 &	61.16 &	69.33 &	50.89 &	48.43 &	63.16 &	71.57 &	70.53 &	69.14 &	67.93 &	55.07 &	65.24 \\
\bottomrule
\end{tabular}
\caption{Ablation results on the XNLI dataset using Greek and Urdu as the auxiliary languages~(averaged over five runs). Results show that ablating the layer-wise update rate or cross-lingual meta-training degrades accuracy on all target languages.}
\label{tab:ablation}
\end{table*}

\subsection{Model and Training Configurations}
Our model is based on the multilingual BERT (mBERT)~\citep{Devlin2019} implemented in GluonNLP~\citep{GluonNLP2020}. As in previous work~\citep{Devlin2019,WuD2019}, we tokenize the input sentences using WordPiece, concatenate them, feed the sequence to BERT, and use the hidden representation of the first token~($[CLS]$) for classification. The final output is computed by applying a linear projection and a softmax layer to the hidden representation. We use a dropout rate of~$0.1$ on the final encoder layer and fix the embedding layer during fine-tuning. Following \citet{NooralahzadehBBA2020}, we fine-tune mBERT by 
\begin{inparaenum}[1)]
    \item fine-tune mBERT on the English data for one epoch to get initial model parameters, and
    \item continue fine-tuning the model on the other source languages for two epochs.
\end{inparaenum}
We compare using the standard optimizer~(fine-tuning baseline) and our meta-optimizer for Step~2.
We use Adam optimizer~\citep{Kingma2015} with a learning rate of~$2 \times 10^{-5}$,~$\beta_1 = 0.9$, and~$\beta_2 = 0.999$ as the standard optimizer and base optimizer in our meta-optimizer. To train our meta-optimizer, we use Adam with a learning rate of~$0.05$ for~$N = 10$ epochs with~$L = 15$ training batches per iteration~(Algorithm~\ref{alg:train}). 
Different from \citet{NooralahzadehBBA2020} who select the auxiliary languages for each target language that lead to the best transfer results, we simulate a more realistic scenario where only a limited set of auxiliary languages is available. We choose two distant auxiliary languages \---\ Greek (Hellenic branch of the Indo-European language family) and Urdu (Indo-Aryan branch of the Indo-European language family) \---\ and evaluate the transfer performance on the other languages. 

\subsection{Main Results} 
\looseness=-1
As shown in Table~\ref{tab:xnli_result}, we compare our meta-learning approach with the fine-tuning baseline and the zero-shot transfer results reported in prior work that uses mBERT. 
Our approach outperforms the fine-tuning methods in \citet{Devlin2019} by~1.6--8.5\%. Compared with the best fine-tuning method in \citet{WuD2019} which freezes a selected subset of mBERT layers during fine-tuning, our approach achieves~+0.4\% higher accuracy on average. We compare our approach with a strong fine-tuning baseline which achieves competitive accuracy scores to the best X-MAML results~\citep{NooralahzadehBBA2020} using a single auxiliary language, even though we limit our choice of the auxiliary language to Greek and Urdu, while \citet{NooralahzadehBBA2020} select the best auxiliary language among all languages except for the target one. Overall, our approach outperforms the strong fine-tuning baseline on~10 out of~14 languages and by~+0.2\% accuracy on average.

Our approach brings larger gains when using two auxiliary languages \---\ it outperforms the fine-tuning baseline on all languages and improves the average accuracy by~+0.6\%. This suggests that our meta-learning approach is more effective when transferring from multiple source languages.\footnote{Using two auxiliary languages improves over one auxiliary language the most on lower-resource languages in mBERT pre-training~(such as Turkish and Hindi), but does not bring gains or even hurts on high-resource languages~(such as French and German). This is consistent with the findings in prior work that the choice of the auxiliary languages is crucial in cross-lingual transfer~\citep{Lin2019}. We leave further investigation on its impact on our meta-learning approach for future work.}

\subsection{Ablation Study} 
\looseness=-1
Our approach is different from \citet{Andrychowicz2016} in that 
\begin{inparaenum}[1)]
    \item it adopts layer-wise update rates while the meta-parameters are shared across all model parameters in \citet{Andrychowicz2016}, and
    \item it trains the meta-parameters in a cross-lingual setting while \citet{Andrychowicz2016} is designated to few-shot learning.
\end{inparaenum}
We conduct ablation experiments on XNLI using Greek and Urdu as the auxiliary languages to understand how they contribute to the model performance.

\paragraph{Impact of Layer-Wise Update Rate} 
We compare our approach with its variant that replaces the layer-wise update rate with one update rate for all layers. Table~\ref{tab:ablation} shows that our approach significantly outperforms this variant on all target languages with an average margin of~2.0\%. This suggests that layer-wise update rate contributes greatly to the effectiveness of our approach.

\paragraph{Impact of Cross-Lingual Meta-Training} 
\looseness=-1
We measure the impact of cross-lingual meta-training by replacing the cross-lingual meta-training in our approach with a joint training of the layer-wise update rate and model parameters. As shown in Table~\ref{tab:ablation}, ablating the cross-lingual meta-training degrades accuracy significantly on all target languages by 1.4\% on average, which shows that our cross-lingual meta-training strategy is beneficial.
\section{Related Work}

\subsection{Cross-lingual Transfer Learning}
The idea of cross-lingual transfer is to use the annotated data in the source languages to improve the task performance on the target language with minimal or even zero target labeled data (aka zero-shot). 
There is a large body of work on using external cross-lingual resources such as bilingual word dictionaries~\citep{Prettenhofer2010,SchusterRBG2019,LiuWLXF2020}, MT systems~\citep{Wan2009}, or parallel corpora~\citep{Eriguchi2018,Yu2018,Singla2018,Conneau2018} to bridge the gap between the source and target languages. Recent advances in unsupervised cross-lingual representations have paved the road for transfer learning without cross-lingual resources~\citep{Yang2017,Chen2018,Schuster2019}.
Our work builds on \citet{Mulcaire2019,LampleC2019,Pires2019} who show that language models trained on monolingual text from multiple languages provide powerful multilingual representations that generalize across languages. Recent work has shown that more advanced techniques such as freezing the model's bottom layers~\citep{WuD2019} or continual learning~\citep{LiuWMF2020} can further boost the cross-lingual performance on downstream tasks. In this paper, we explore meta-learning to softly select the layers to freeze during fine-tuning.

\subsection{Meta Learning}
A typical meta-learning algorithm consists of two loops of training:~1) an \textit{inner loop} where the learner model is trained, and~2) an \textit{outer loop} where, given a meta-objective, we optimize a set of meta-parameters which controls  aspects of the learning process in the inner loop. The goal is to find the optimal meta-parameters such that the inner loop performs well on the meta-objective. Existing meta-learning approaches differ in the choice of meta-parameters to be optimized and the meta-objective. Depending on the choice of meta-parameters, existing work can be divided into four categories: (a) neural architecture search~\citep{Stanley2002,Zoph2016,Baker2016,Real2017,Zoph2018}; (b) metric-based~\citep{Koch2015,Vinyals2016}; (c) model-agnostic (MAML)~\citep{Finn2017,Ravi2016}; (d) model-based (learning update rules)~\citep{Schmidhuber1987,Hochreiter2001,Maclaurin2015,Li2017}.

\looseness=-1
In this paper, we focus on model-based meta-learning for zero-shot cross-lingual transfer. Early work introduces a type of networks that can update their own weights~\citep{Schmidhuber1987,Schmidhuber1992,Schmidhuber1993}. 
More recently, \citet{Andrychowicz2016} propose to model gradient-based update rules using an RNN and optimize it with gradient descent. However, as \citet{Wichrowska2017} point out, the RNN-based meta-optimizers fail to make progress when run for large numbers of steps. They address the issue by incorporating features motivated by the standard optimizers into the meta-optimizer. We instead base our meta-optimizer on a standard optmizer like Adam so that it generalizes better to large-scale training.

\looseness=-1
Meta-learning has been previously applied to few-shot cross-lingual named entity recognition~\citep{Wu2019}, low-resource machine translation~\citep{Gu2018}, and improving cross-domain generalization for semantic parsing~\citep{Wang2021}. For zero-shot cross-lingual transfer, \citet{NooralahzadehBBA2020} introduce an optimization-based meta-learning algorithm called X-MAML which meta-learns the initial model parameters on supervised data from low-resource languages. By contrast, our meta-learning algorithm requires much less meta-parameters and is thus simpler than X-MAML. \citet{BansalJM2020} show that MAML combined with meta-learning for learning rates improves few-shot learning. Different from their approach which learns layer-wise learning rates only for task-specific layers specified as a hyper-parameter as part of the MAML algorithm, our approach learns layer-wise learning rates for all layers, and we show the effectiveness of our approach without being used with MAML on zero-shot cross-lingual transfer.

\section{Conclusion}
\looseness=-1
We propose a novel meta-optimizer that learns to soft-select which layers to freeze when fine-tuning a pretrained language model (mBERT) for zero-shot cross-lingual transfer. Our meta-optimizer learns the update rate for each layer by simulating the zero-shot transfer scenario where the model fine-tuned on the source languages is tested on an unseen language. Experiments show that our approach outperforms the simple fine-tuning baseline and the X-MAML algorithm on cross-lingual natural language inference.

\bibliography{acl2021}
\bibliographystyle{acl_natbib}

\end{document}